\documentclass{article}

    \usepackage[preprint]{neurips_2022}

\usepackage[utf8]{inputenc} 
\usepackage[T1]{fontenc}    
\usepackage{hyperref}       
\usepackage{url}            
\usepackage{booktabs}       
\usepackage{amsfonts}       
\usepackage{nicefrac}       
\usepackage{microtype}      
\usepackage{xcolor}         
\usepackage{graphicx}
\usepackage{natbib}

\title{Survey on Self-Supervised Multimodal Representation Learning and Foundation Models}

\author{%
  Sushil Thapa\thanks{Work done as a part of coursework "CSE-585: Graduate Seminar" under Prof. Dr. Clinton L. Jeffery.} \\
  Department of Computer Science and Engineering\\
  New Mexico Tech\\
  \texttt{sushil@cs.nmt.edu} \\
}

\begin{document}

\maketitle

\begin{abstract}
  Deep learning has been the subject of growing interest in recent years. Specifically, a specific type called Multimodal learning has shown great promise for solving a wide range of problems in domains such as language, vision, audio, etc. One promising research direction to improve this further has been learning rich and robust low-dimensional data representation of the high-dimensional world with the help of large-scale datasets present on the internet. Because of its potential to avoid the cost of annotating large-scale datasets, self-supervised learning has been the de facto standard for this task in recent years. This paper summarizes some of the landmark research papers that are directly or indirectly responsible to build the foundation of multimodal self-supervised learning of representation today. The paper goes over the development of representation learning over the last few years for each modality and how they were combined to get a multimodal agent later. 
\end{abstract}

\section{Introduction}
Deep learning has advanced to the point that it is now one of the most important components of most intelligent systems. Deep neural networks (DNNs) are a compelling approach in computer vision (CV) tasks and natural language processing (NLP) tasks due to their ability to learn rich patterns from the data available today.
However, because of the high-cost requirements to annotate datasets, the supervised approach of learning features from labeled data has practically achieved saturation. To avoid this, researchers nowadays have started to learn the supervisory signals without explicit supervision from humans. Since the model self-learns the supervision from the data itself, it is called self-supervised learning which is different from Supervised learning where we explicitly annotate supervision, Unsupervised where we have no supervision whatsoever, and Reinforcement learning where we get the rewards from the environment for our steps. Such models use billions of public images, texts, or other modality datasets to learn features that help to have a fundamental understanding of the world around us. It started out as using neural networks to learn language models\citep{neuralLM}and learning distributed representations of words \citep{word2vec}. Once the attention came in place\citep{bahdanau2014neural} \citep{transformer}, NLP has had breakthrough progress on various NLP tasks\citep{bert}\citep{floridi2020gpt}\citep{lan2019albert}. Following the advances in NLP, researchers also tried exploring similar problem formulation on other modalities like images \citep{dosovitskiy2020image}\citep{wu2021cvt}, audio \citep{chi2021audio}\citep{chuang2019speechbert} and the combination of at least two of those modalities\citep{akbari2021vatt}\citep{su2019vlbert}\citep{wang2020linformer}. 

This paper surveys the development of each of such modalities and the progress of combining those modalities into a single system. The goal is to discuss the emerging self-supervised techniques for such modalities and understand the motivation of multimodal fusion to build a model that can perceive the world much like how our senses do. There are hundreds of papers that investigate this, but we filtered them by choosing only the influential papers with good reputations and progress over previous systems. This is, as per our knowledge, the most recent and most comprehensive survey that focuses on this area of research.

\section{Research Questions}
The paper initially establishes how each modality representation improved through a literature review. It discusses language, vision, audio, and robotics applications separately. It then follows up with the other type of research where they combine such modalities in one way or the other. In summary, it asks the following research questions focusing on the methodologies of learning such representations. 
\begin{itemize}
    \item What was the motivation for learning self-supervised representations?
    \item How the different techniques could be applied for learning representations for multiple types of modalities?
    \item How can the multiple separate modalities be combined to get more effective AI agents? What do we gain by combining them?
\end{itemize}

\section{Related Work}
This section focuses on the progress in the individual modalities separately. The success of each of these modalities is the foundation of the current success in multimodal systems. 
\subsection{Language}

\subsubsection{Language Model and Embedding}
\begin{figure}
	\centering
	\includegraphics[width=\linewidth]{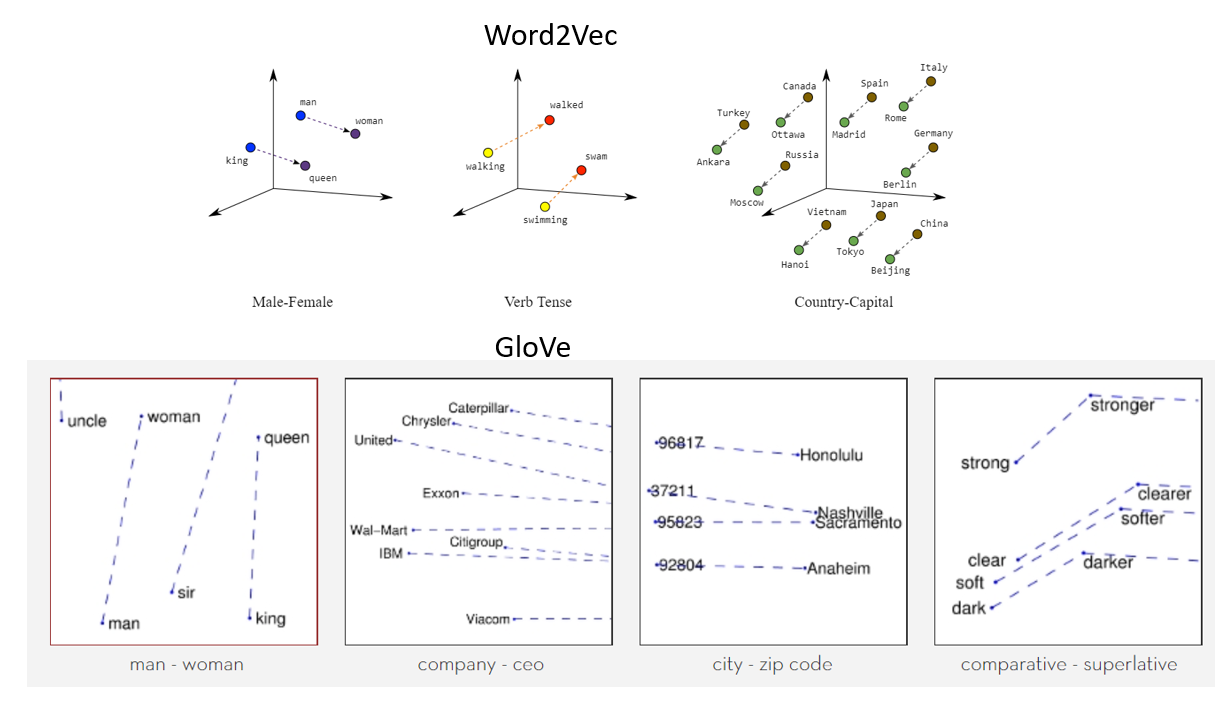}
	\caption{Visualizations of Learned Embedding\citep{word2vec}}
	\label{fig:embeddings}
\end{figure}

With the motivation of learning statistical models for defining languages, this paper proposes a method to learn the statistical joint probability distribution function of sequences of words that come together in sentences. This paper eliminates the curse of dimensionality by learning a distributed representation for words by training with neighboring words in sentences. 
In this statistical model\citep{neuralLM}, the words and language can be represented by the conditional probability of the next word given all previous ones, such that
\begin{displaymath}
    \hat{P}(w_1^T)=\prod_{t=1}^{T}\hat{P}(w_t | w_1^{t-1})
\end{displaymath}
where $w_t$ is the $t$-th word and writing sub-sequence\\ $w_i^j=(w_i, w_{i+1},...,w_{j-1}, w_j)$. However, analyzing all the previous occurred can be compute-intensive and slow, so they used a setup that would only look at past $n$ words by building $n-grams$ to learn the context. Larger the value of $n$, the bigger context they would learn from it. Now the right expression approximately becomes: 
\begin{displaymath}
   \hat{P}(w_t | w_1^{t-1}) \approx \hat{P}(w_t | w_{\textcolor{red}{t-n+1}}^{t-1})
\end{displaymath}
Such language models were great to get a sense of what is the next possible word given a sequence of previous words. However, instead of learning a model with previous words, later works\citep{word2vec} \citep{pennington2014glove} focused more on building a big embedding representation of words based on the company they keep. They used contexts to predict the associated words and also learn language models. They also used a simple linear model to go through billions of words within a day. Interestingly, when we looked at the embedding it learned through such a setup, we could get the interpretable association of analogies as shown in Figure \ref{fig:embeddings}.

\subsubsection{Attention}
\begin{figure}[!htb]
	\centering
	\includegraphics[width=\linewidth]{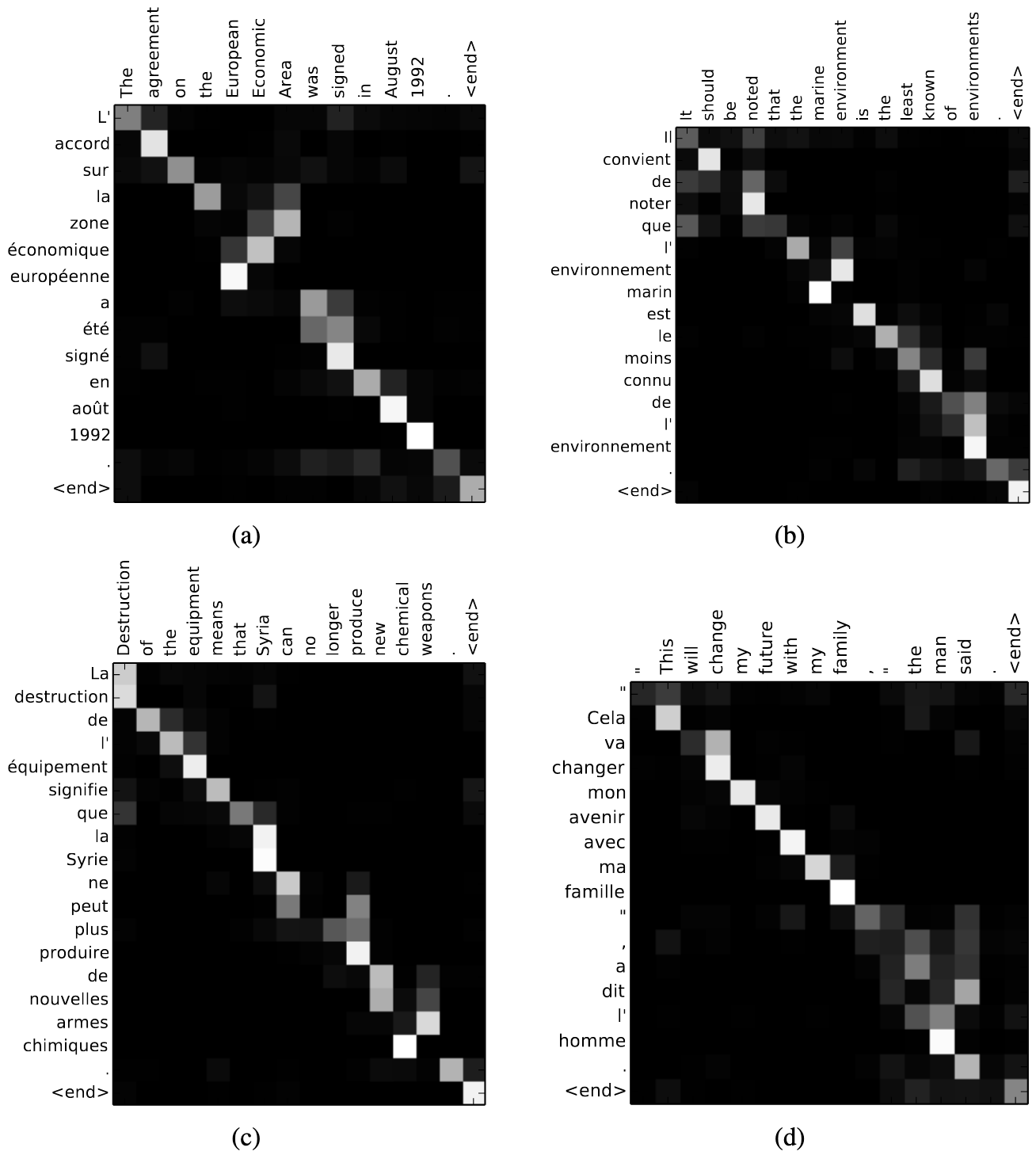}
	\caption{Four sample alignments. The x-axis and y-axis of each plot
correspond to the words in the source sentence (English) and the generated translation (French),
respectively. Each pixel shows the weight of the match of annotation of the $j-th$ source word for the $i-th$ target word\citep{bahdanau2014neural}}
	\label{fig:attention}
\end{figure}

\begin{figure}[!htb]
	\centering
	\includegraphics[width=\linewidth]{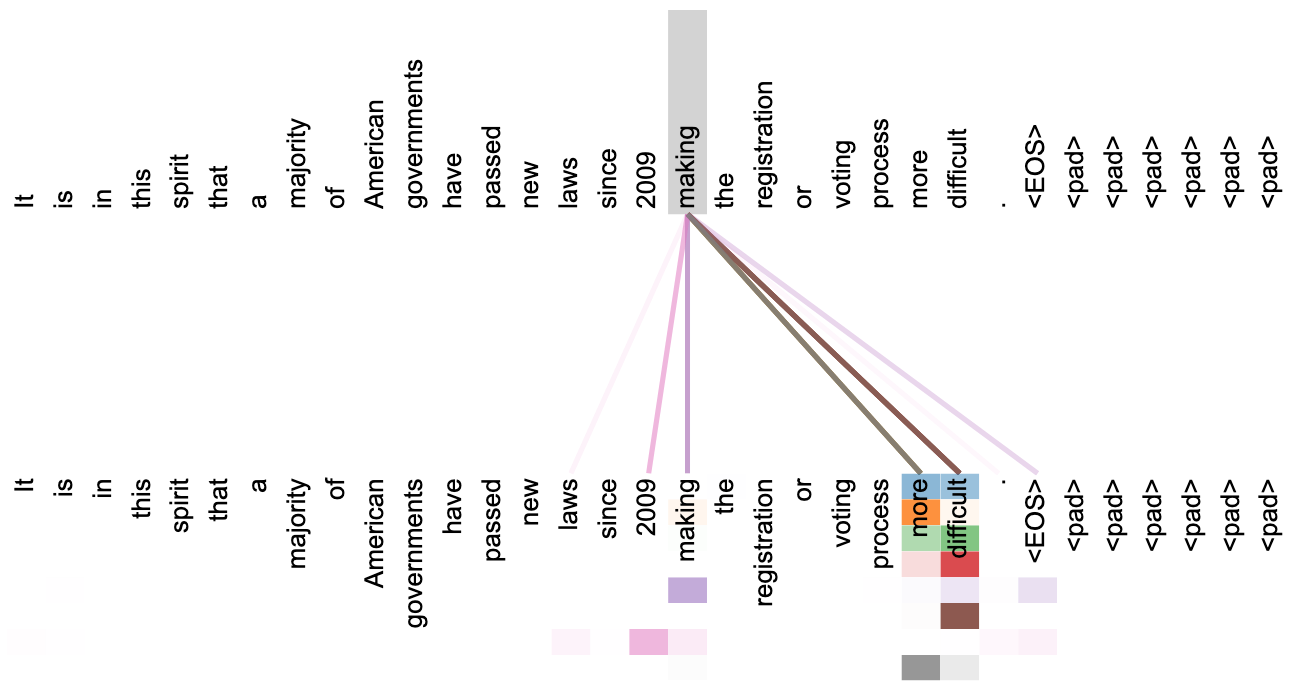}
	\caption{An example of the self-attention mechanism following long-distance dependencies in the
encoder self-attention. Many of the attention heads attend to a distant dependency of
the verb ‘making’, completing the phrase ‘making...more difficult’\citep{transformer}.}
	\label{fig:selfattention}
\end{figure}

This section focuses on the origin and motivation for developing attention mechanisms in neural networks. For mapping sequences of inputs to output sequences of output like in Machine Translation, this work\citep{sutskever2014sequence} proposed modeling/encoding the input sequence for eg. an English sentence to a fixed-sized vector representation and later decoding it to give a sequence of output for eg. German in English to German language translation. This was effective but encoding a variable-length sequence into a fixed-length vector wasn't an intuitive thing to do. Now, instead of trying to decode from a fixed vector, this work\citep{bahdanau2014neural} introduced attention where the decoder can essentially attend/focus on a specific region of the source sequence directly. This allows the model to learn the one-one mapping of relevant words and as shown in Figure \ref{fig:attention} we could also interpret the attention weights of words across a whole sentence.
Attention was so powerful and revolutionary at that time, this work\citep{transformer} essentially proposes to remove the whole sequence/recurrence bit from the model. They fundamentally just used a type of attention to process the whole sequence in parallel with the help of positional encoding. This model was called Transformers\citep{transformer} that would employ a self-attention technique that allows the model to learn representations by looking at the input sequence itself as shown in Figure \ref{fig:selfattention}. This Transformer model\citep{transformer} actually revolutionized the Machine learning research as we know it today.

\subsubsection{BERT Family}
With the help of self-attention in transformers, the encoders learn rich representations of the words with the help of context. In following parallel work\citep{peters2018deepelmo}, they tried making the embedding context-dependent as well with the help of bi-directional LSTM\citep{hochreiter1997lstm}. Likewise, works like this\citep{radford2018gpt} focused more on the idea of pretraining and learning representation for language understanding with Transformers. Later, to condition the context not only on each side, BERT\citep{bert} learns representation looking both forward and backward inspired by Cloze procedure\citep{taylor1953cloze}. To learn the word-words interaction, they tried masking a word and using context around it to predict the masked word. The idea is if the model is able to predict it successfully, it really learned about the syntax and semantics  of the words. With the help of dummy tasks like if the two sentences given to it are contiguous(Next Sentence Prediction task) or not, it learns the more broad sentence level representation. This was a major breakthrough in the NLP community. The fully self-supervised representations that it learns were so rich that when applied to 11 downstream tasks ranging from Machine translation, Question Answering, etc. Later vanilla Bert\citep{bert} is again scaled up for robustness\citep{liu1907roberta}, added recurrence\citep{dai1901transformer}, replaced Masked with Permutation Language modeling\citep{yang1906qv} among other things to achieve a further gain in the performance. 

\begin{figure}[!htb]
	\centering
	\includegraphics[width=\linewidth]{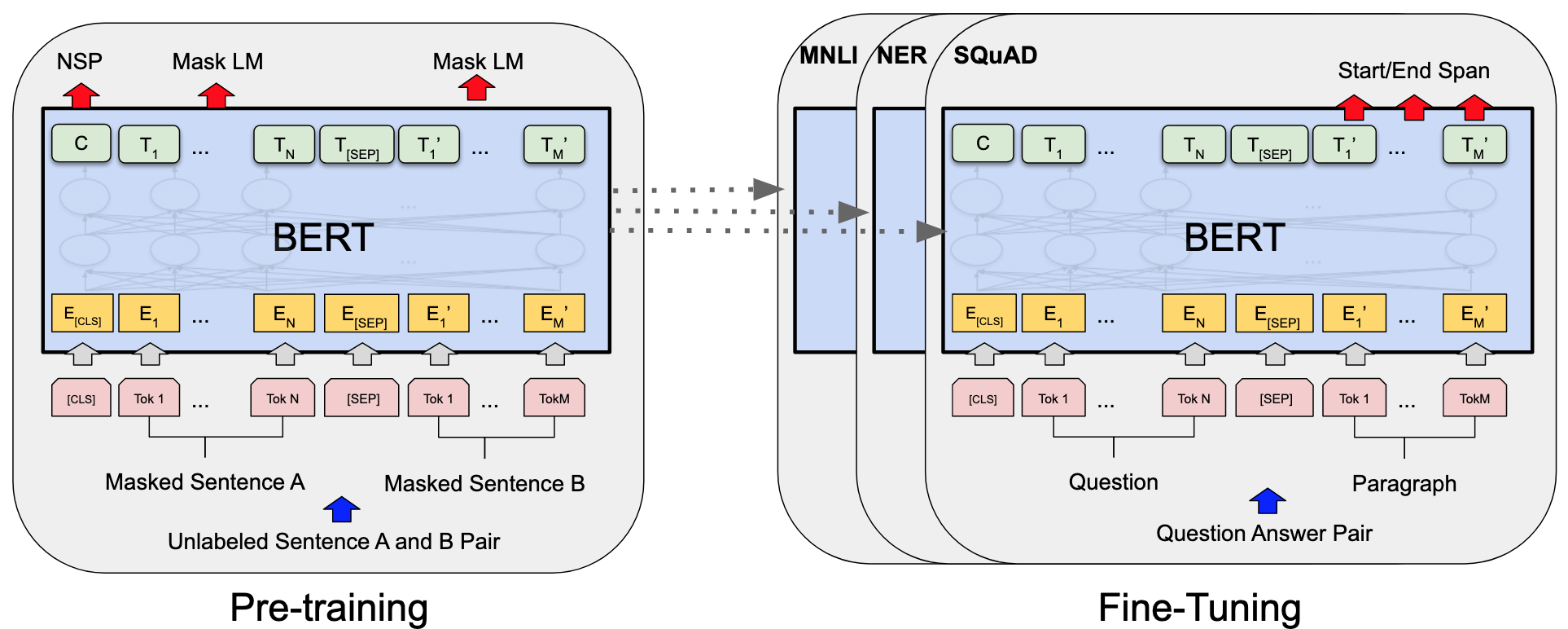}
	\caption{Overall pre-training and fine-tuning procedures for BERT. Apart from output layers, the same architectures are used in both pre-training and fine-tuning. The same pre-trained model parameters are used to initialize models for different downstream tasks.\citep{bert}.}
	\label{fig:bert}
\end{figure}

Many of the state-of-the-art models that we see today are in one form or another other a variant of BERT and the transformers.

\subsection{Image}
Just like NLP, Vision has also been adapting self-supervised techniques to learn the representations. Self-supervision in image representation learning uses image-related pretext tasks that work similarly to the mask Language Model in BERT. It helps learn the actual representation without explicit supervision. 

\subsubsection{Pretext tasks}
\begin{figure}[!htb]
	\centering
	\includegraphics[width=0.7\linewidth]{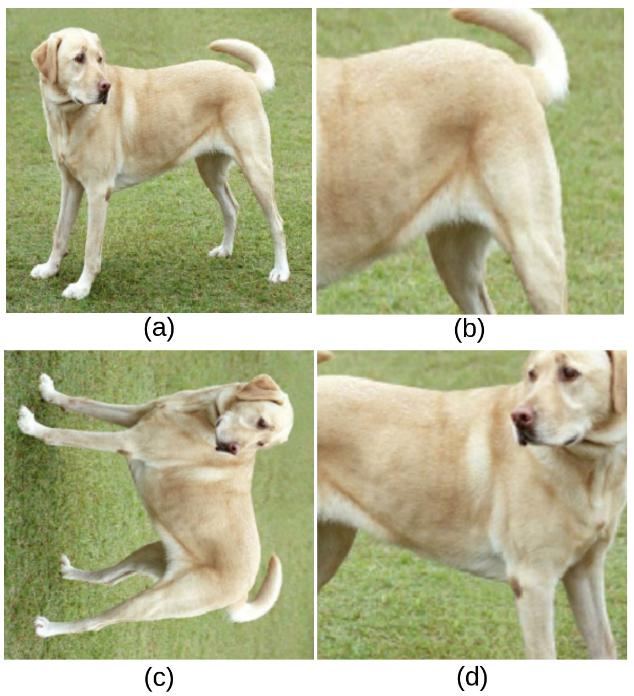}
	\caption{Geometric Transformation as pretext task \citep{simclr}. (a) Original (b) Crop and Resize (c) Rotate(90◦, 180◦, 270◦) (d) crop, resize, flip.}
	\label{fig:aug}
\end{figure}

These tasks generate pseudo labels for a dummy prediction problem. The learned model from this has features that are much richer and more condensed than the high-dimensional images. Those features can then be used for different downstream tasks like classification, detection, etc. Pretext tasks are generally not associated with images but can be applied to other domains as well. For the pretext task, in this case, the original dataset is an anchor, the augmented version usually defines the positive samples and other images in the dataset as the negative samples. Generally, Pretext tasks can be divided into four main categories\citep{Jaiswal_2020contrastive}:
\begin{itemize}
    \item visual transformation: generally basic adjustments that alter visual pixels/visual aspects like blurring, noise, distortion, etc. 
    \item geometric transformation: spatial transformations such as scaling, rotation, cropping, etc. as shown in Figure \ref{fig:aug}
    \item context-based tasks: Manipulates context the data is in. For eg. Jigsaw puzzles of image crops, future or missing frame prediction from videos, etc. 
    \item view prediction tasks: The same object with different views are positive sample in this case.
\end{itemize}

\subsubsection{Contrastive learning methods}
\begin{figure}[!htb] 
	\centering
	\includegraphics[width=\linewidth]{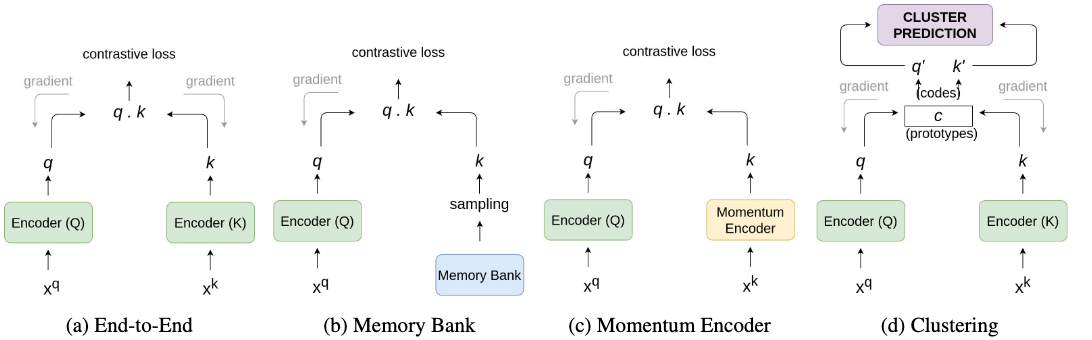}
	\caption{Different architecture pipelines for Contrastive Learning: (a) End-to-End training of two encoders where one generates representation for positive samples and the other for negative samples (b) Using a memory bank to store and retrieve encoding of negative samples (c) Using a momentum encoder which acts as a dynamic dictionary lookup for encoding of negative samples during training (d) Implementing a clustering mechanism by using swapped prediction of the obtained representations from both the encoders using end-to-end architecture\citep{Jaiswal_2020contrastive}.}
	\label{fig:contrastive_pipeline}
\end{figure}

\begin{figure}[!htb]
	\centering
	\includegraphics[width=\linewidth]{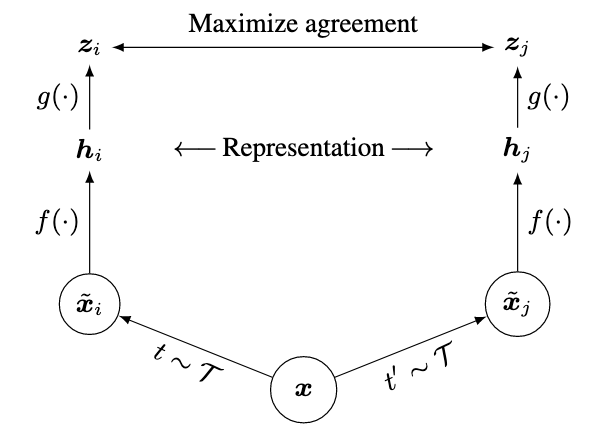}
	\caption{A simple framework for contrastive learning of visual representations. Two separate data augmentation operators are sampled from the same family of augmentations ($t \sim T$ and $t\;' \sim T$) and applied to each data example to obtain two correlated views. \citep{simclr}.}
	\label{fig:simclr}
\end{figure}

    This is the self-supervised technique to learn the visual features by contrasting different forms of images to other images. As shown in Figure \ref{fig:contrastive_pipeline} there can be different contrastive approaches. In general end-to-end frameworks like simCLR\citep{simclr} just computes contrastive loss across augmented views of an anchor image to other images as shown in Figure \ref{fig:simclr}. Likewise, by adding momentum\citep{moco} it was able to train bigger batches of negative samples. In summary, the augmented version of anchor $z_i$ maximizes the agreement whereas it is minimized with a different image.
    
\subsubsection{Clustering methods}
\begin{figure}[!htb]
	\centering
	\includegraphics[width=\linewidth]{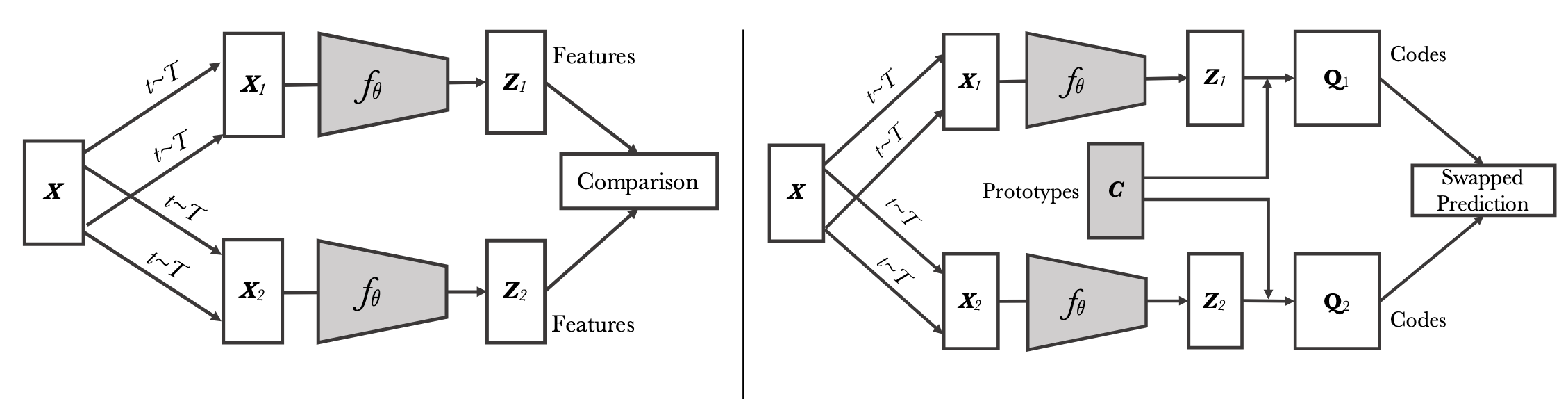}
	\caption{Contrastive instance learning (left) vs. SwAV (right). In contrastive learning methods applied to instance classification, the features from different transformations of the same images are compared directly to each other. In SwAV, we first obtain “codes” by assigning features to prototype vectors. We then solve a “swapped” prediction problem wherein the codes obtained from one data-augmented view are predicted using the other view. Thus, SwAV does not directly compare image features. Prototype vectors are learned along with the ConvNet parameters by back-propagation. \citep{caron2020unsupervisedswav}}
	\label{fig:swav}
\end{figure}

This method tries to cluster the augmented version of images together in an unsupervised fashion like in DeepCluster\citep{caron2018deepcluster}. It has a specific number of clusters that need to cluster all the available datasets. Whereas in the other types of methods like SwAV\citep{caron2020unsupervisedswav}, there is a prototypical representation of different views of the dataset like in Figure \ref{fig:swav}, it is then clustered and the similar vectors are pushed to the similar cluster which doesn't have to be online which is much easier to manage. 

\subsection{Audio}
\begin{figure}[!htb]
	\centering
	\includegraphics[width=\linewidth]{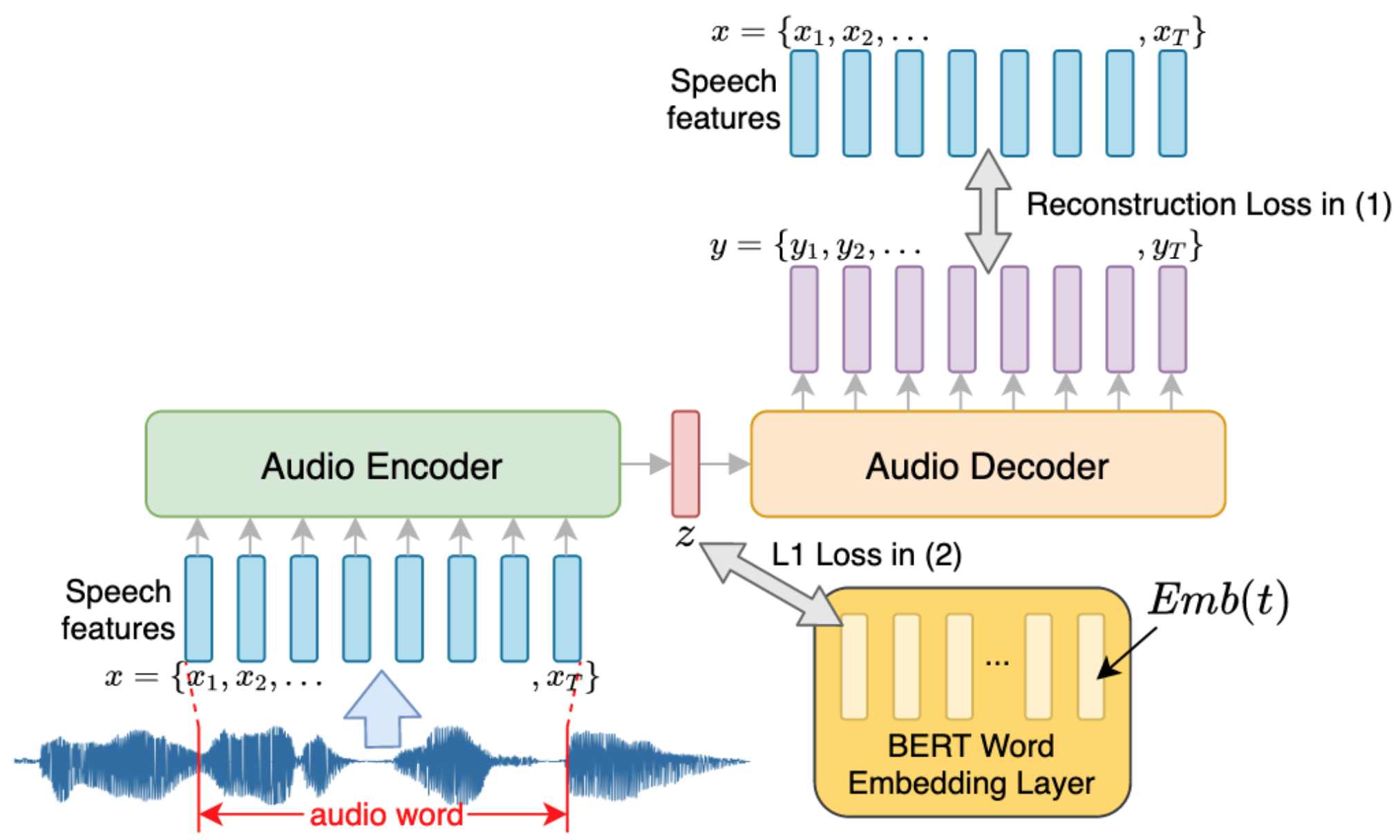}
	\caption{Training procedure for the Initial Phonetic-Semantic Joint Embedding. After training, the encoded vector (z in red) obtained here is used to train the SpeechBERT\citep{chuang2019speechbert}}
	\label{fig:speechbert}
\end{figure}

audiobert, \citep{lakhotia2021generative}\
Just like images and language, Speech modality has also seen some developments in learning speech representations with the help of BERT and Transformers. Digital speech is just a waveform of signals which can be encoded to speech features with the help of a spectrogram. Those speech features are then fed into an audio encoder which learns an embedding $z$. The decoder then takes $z$ and generates the speech features back as shown in Figure \ref{fig:speechbert}. With the help of reconstruction loss, it learns about the reconstruction of the features which ultimately leads to the learning of rich speech representations. This is a type of generative self-supervised learning. 

\subsection{Robotics}
\begin{figure}[!htbp]
	\centering
	\includegraphics[width=\linewidth]{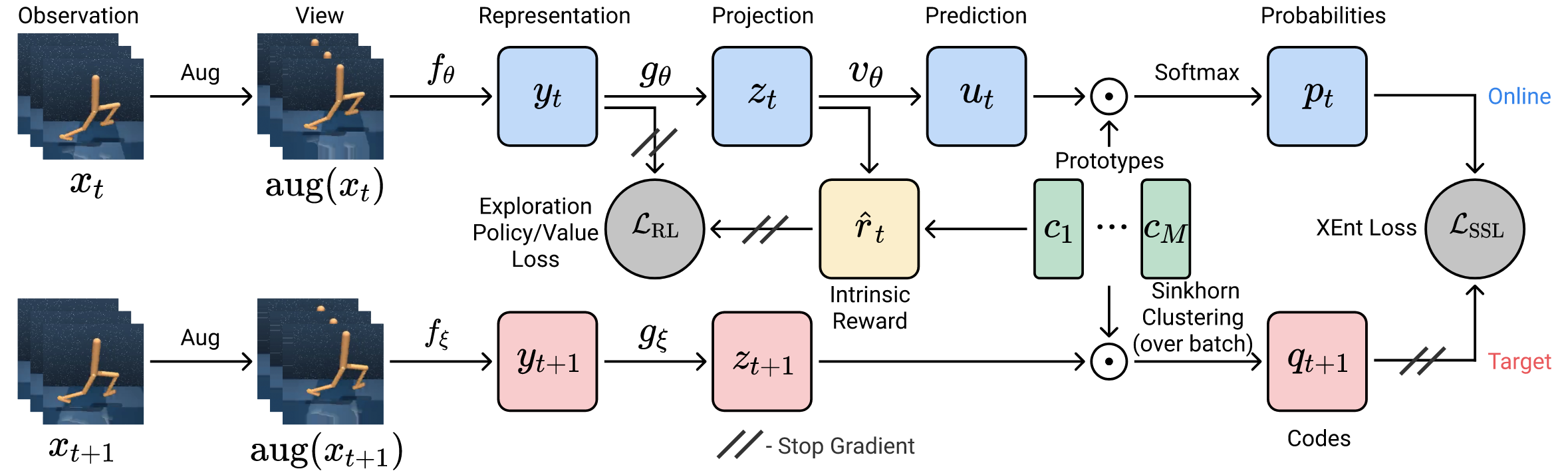}
	\caption{Proto-RL proposes a self-supervised scheme that learns to encode high-dimensional image observations $x_t$, $x_{t+1}$, using an encoder $f\theta$ along with a set of prototypes ${c_i}^M_{i=1}$ that defines the basis of the latent space\citep{yarats2021reinforcementprotorl}}
	\label{fig:protorl}
\end{figure}

Inspired by the recent advances in images like MoCo\citep{moco}, SwAV\citep{caron2020unsupervisedswav}, Reinforcement learning and robotics have also started embracing them to provide auxiliary rewards in case of sparse rewards from the main RL environment. Since these auxiliary rewards are generated self-supervised way, this helps build a powerful sample efficient Reinforcement learning agent. Figure \ref{fig:protorl} shows one such agent. This work is based on the SwAV\citep{caron2020unsupervisedswav} paper but applied to frames of RL environments like games and simulated tasks. 

\subsection{Multimodal}
Till now, we discussed how self-supervised pre-training and fine-tuning have been helpful in different modalities of datasets separately. In this section, We will discuss the combinations of two or more modalities to solve a multimodal task like image captioning, visual question answering, music generation, etc. 

\subsubsection{Visio-linguistic models}
\begin{figure}[!htbp]
	\centering
	\includegraphics[width=\linewidth]{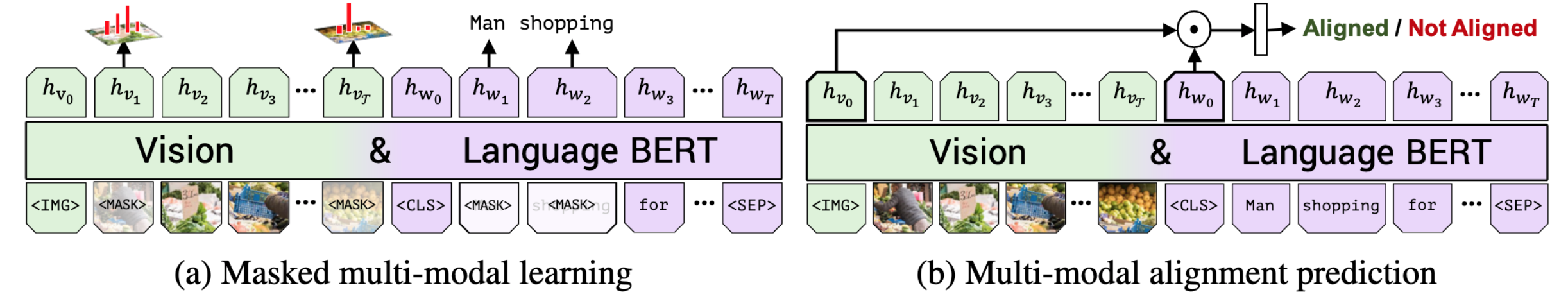}
	\caption{Vilbert training task. In masked multi-modal learning, the model must reconstruct image region categories or words for masked inputs given the observed inputs. In multi-modal alignment prediction, the model must predict whether or not the caption describes the image content.\citep{vilbert}}
	\label{fig:vilbert}
\end{figure}

This is probably the most famous family of multimodal tasks. Due to the abundance of text and image data sources from news portals, social media, etc. We have seen tremendous progress in these types of models. To name a few, these models \citep{vilbert} \citep{visualbert} \citep{su2019vlbert} encode respective modalities with separate encoders and then fuse them together to learn a cross-modal embedding that can learn both visual and textual features. The dummy task here is inherited from the Bert\citep{bert} itself but slightly modified to take in both visual and textual representations. As shown in Figure \ref{fig:vilbert}, The mask language model becomes a masked multimodal learning system where it masks a part of input in one modality and tries to predict it from the remaining part as well as the other modality dataset. Similarly, the next sentence prediction task is also modified to predict whether or not the encoding of one modality matches the other modality. To provide a pseudo-labeled alignment dataset, billions of images with their alt-text/captions from the web. 
\
CLIP\citep{radford2021learning}

\subsubsection{Adding other modalities} 
In addition to these types of models, recently a much bigger model has been developed for supporting as many modalities as possible to learn a shared representation. For example, the Perceiver model\citep{jaegle2021perceiver}, does not make any domain-specific assumptions about the input. It can support high dimensional inputs such as images, videos, audio, point clouds, and other multimodal combinations to predict the logits with the help of Transformers. 
\begin{figure}[!htb]
	\centering
	\includegraphics[width=\linewidth]{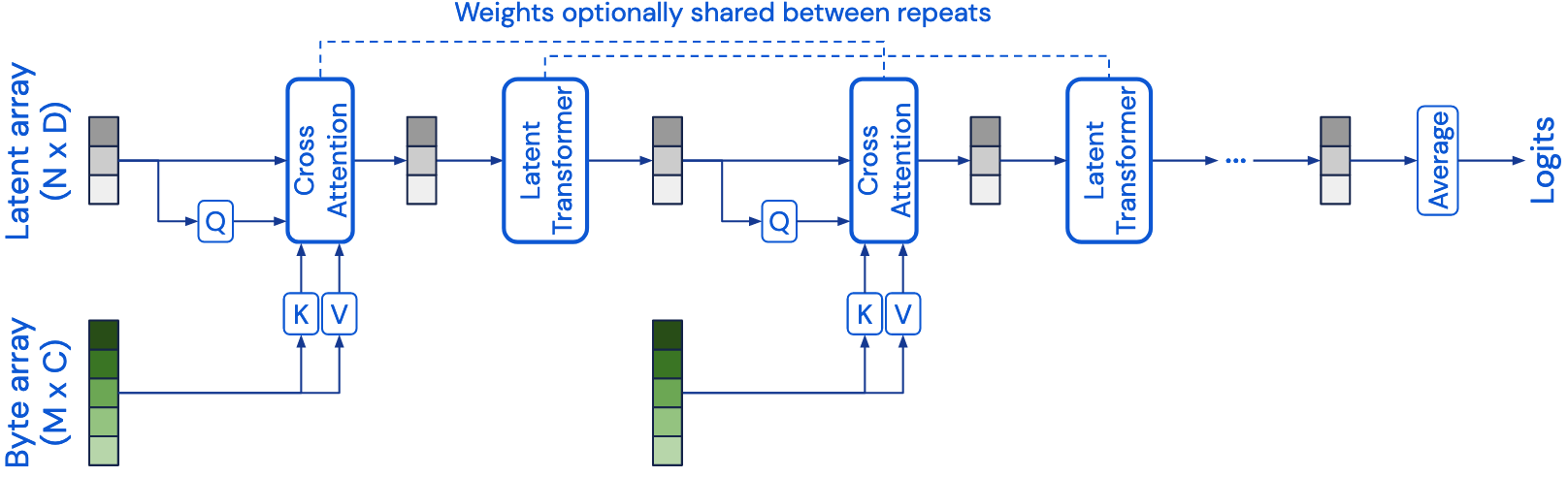}
	\caption{The Perceiver model architecture\citep{jaegle2021perceiver}.}
	\label{fig:perceiver}
\end{figure}

\begin{figure}[!htb]
	\centering
	\includegraphics[width=\linewidth]{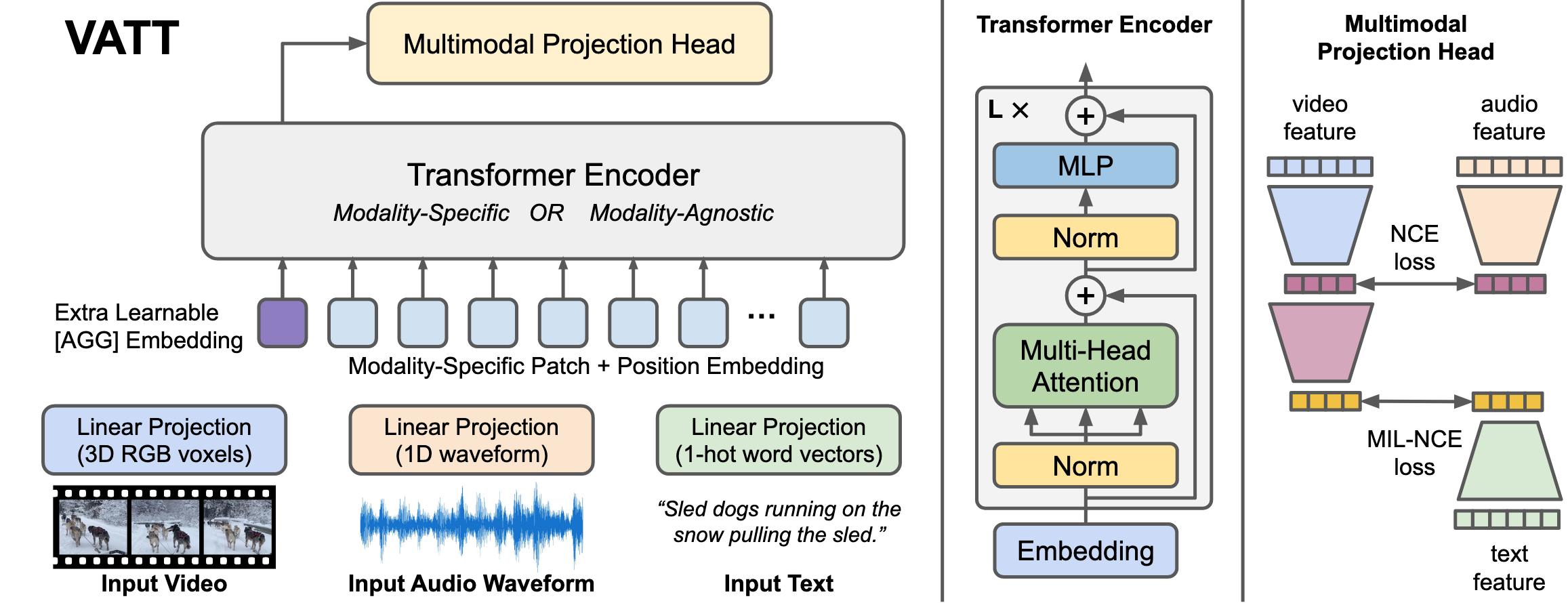}
	\caption{Overview of the VATT architecture and the self-supervised, multimodal learning strategy\citep{akbari2021vatt}}
	\label{fig:vatt}
\end{figure}

Similarly, there is other research where other types of convolution-free transformers-based multimodal models like Video-Audio-Text Transformers(VATT)\citep{akbari2021vatt} are really effective. This type of modality-agnostic single-backbone transformer learns by sharing weights among the three modalities as shown in Figure \ref{fig:vatt}.


\section{Research Opportunities}
Going through this literature, the power of using a carefully designed Machine learning model combined with unlimited data sources can be realized. Even if there has been a lot of progress in this field in recent years, there are a few areas that still need attention from researchers. The main bottleneck in today's multimodal systems seems to be a handcrafted set of augmentations with human understanding. But, If we can somehow get past it to build learnable augmentation systems without human inductive bias and trial-and-error, the progress could be multiplied. Some other direction could be building better pretext tasks that could learn the more rich feature representations. Most of the models that could handle modalities seem to be Transformer based- we think improving the efficiency and setup of transformers could also be really beneficial. Big companies these days are more focused on building large models with large datasets but fail to answer smaller more impactful questions. Model interpretability/explainability and its fairness, and being able to retain the performance in adversarial attack and distribution shifts seem to be some other research directions on these types of models.

\section{Conclusion}
This study extensively discussed a wide range of recent top-\\performing self-supervised representation learning algorithms for vision, NLP tasks, audio, robotics, and the combination of them. From their initial days of learning embedding representations to contrastive features to employing the acquired parameters for downstream tasks, we reviewed each module and modality of such learning history. Multimodal representation has yielded encouraging results in a variety of downstream applications. Finally, this paper discusses some of the open issues with current techniques prevalent in today's learning methods and also their possible ways of improvement. 

\section{Disclaimer}
This paper is supposed to provide a glimpse on the current research trends on multimodal representation learning and is by no means fully comprehensive with everything that is out there. We probably have missed a lot of awesome papers and manuscripts. Please feel free to email me with any recommendations that you think should be included here.  

\bibliography{bibliography}






\end{document}